\documentclass[10pt,twocolumn,letterpaper]{article}

\usepackage{wacv}              

\usepackage{graphicx}
\usepackage{amsmath}
\usepackage{amssymb}
\usepackage{booktabs}
\usepackage{float}

\usepackage[pagebackref,breaklinks,colorlinks]{hyperref}

\usepackage[capitalize]{cleveref}
\crefname{section}{Sec.}{Secs.}
\Crefname{section}{Section}{Sections}
\Crefname{table}{Table}{Tables}
\crefname{table}{Tab.}{Tabs.}

\def\wacvPaperID{3} 
\def\confName{WACV}
\def\confYear{2025}

\begin{document}

\title{Towards long-term player tracking with graph hierarchies and domain-specific features}

\author{Maria Koshkina\\
York University\\
Toronto, Canada\\
{\tt\small koshkina@yorku.ca}
\and
James H. Elder\\
York University\\
Toronto, Canada\\
{\tt\small jelder@yorku.ca}
}
\maketitle

\begin{abstract}
In team sports analytics, long-term player tracking remains a challenging task due to player appearance similarity, occlusion, and dynamic motion patterns. Accurately re-identifying players and reconnecting tracklets after extended absences from the field of view or prolonged occlusions is crucial for robust analysis. We introduce SportsSUSHI, a hierarchical graph-based approach that leverages domain-specific features, including jersey numbers, team IDs, and field coordinates, to enhance tracking accuracy.  SportsSUSHI achieves high performance on the SoccerNet dataset and a newly proposed hockey tracking dataset. Our hockey dataset, recorded using a stationary camera capturing the entire playing surface, contains long sequences and annotations for team IDs and jersey numbers, making it well-suited for evaluating long-term tracking capabilities. The inclusion of domain-specific features in our approach significantly improves association accuracy, as demonstrated in our experiments.  The dataset and code are available at \href{https://github.com/mkoshkina/sports-SUSHI}{https://github.com/mkoshkina/sports-SUSHI}
\end{abstract}

\section{Introduction}
\label{sec:intro}
\begin{figure}[ht]
  \centering
   \includegraphics[width=0.8\linewidth]{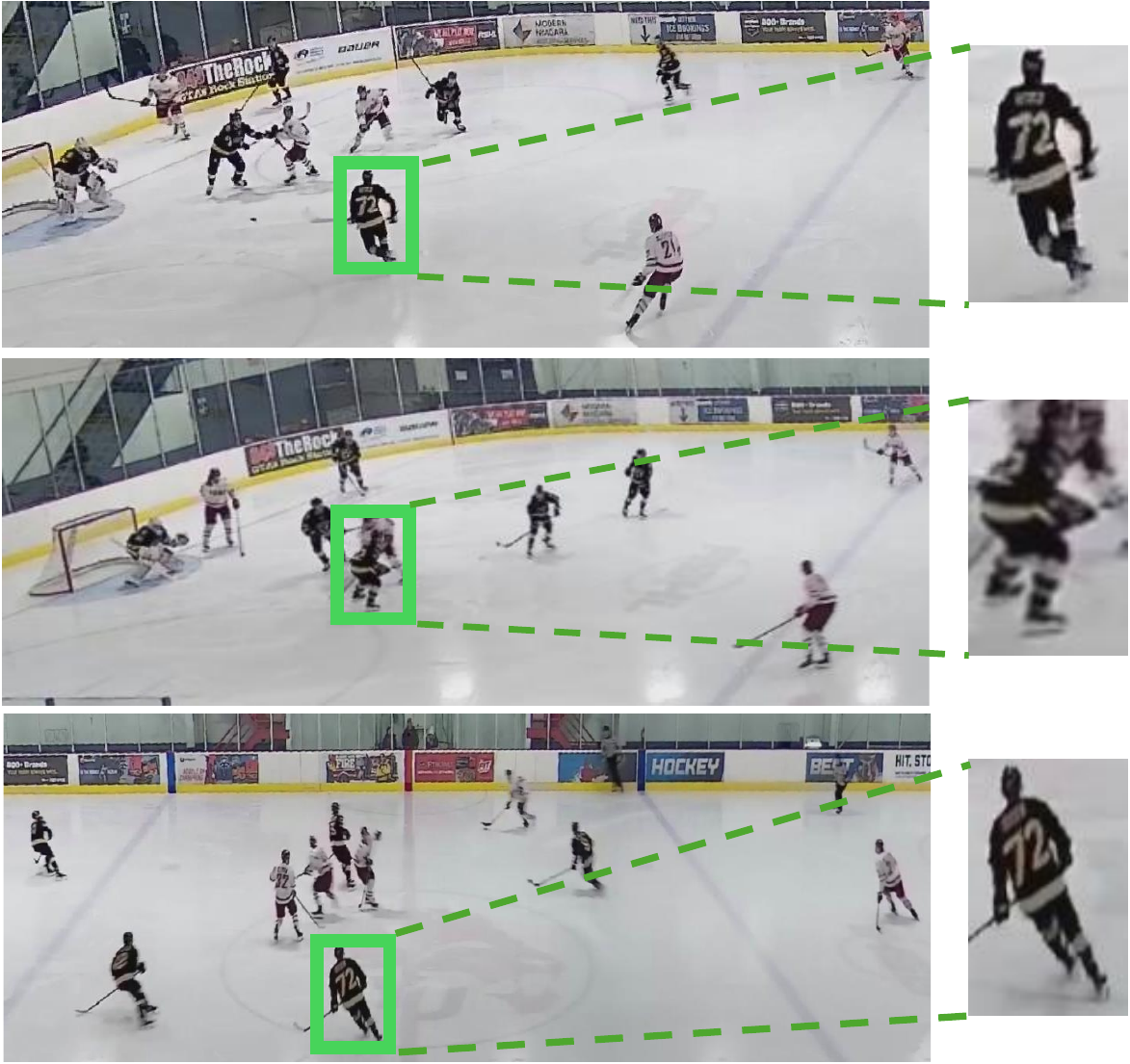}

   \caption{Team uniform and protective gear make players on the same team hard to distinguish. Jersey number is a key feature for player re-identification and long-term player tracking.}
   \label{fig:onecol}
\end{figure}

Sports video understanding and automatic statistics are
important applications of computer vision. They help coaches
improve tactics and run better training, enable efficient game
analysis, and can enrich the sports viewing experience. The
ability to automatically track players during a team game is
essential for game analysis. It involves detecting and following
players during a game. Player tracking is a
specific case of a well-studied topic in computer vision - Multi-Object Tracking (MOT). Typically, MOT is studied in the
context of pedestrian tracking (for example, for surveillance
purposes) and vehicle tracking (for self-driving cars and traffic
analysis). Tracking players during a team sports event shares
the same problem setup and challenges as tracking pedestrians.
Generic MOT tracking algorithms can be applied to tracking
players and will produce reasonable results. However, there
are several important differences between generic MOT
and team sport MOT:
\begin{itemize}
\item Players on the same team often look similar, especially in winter sports like hockey, where gear obscures distinctive features, making traditional re-identification methods less effective.

\item Pedestrian tracking focuses on short-term tracking, while player tracking in sports requires recovering tracks after long occlusions or absences.

\item Players move quickly and often follow non-linear motion patterns.

\item Sports settings involve fewer individuals in view compared to pedestrian scenes.

\item Players can be uniquely identified by their team and jersey numbers.

\end{itemize}

Most recent approaches to player tracking utilize online short-term tracking \cite{huang2023observation, yang2023hard, lv2024diffmot}. Some apply the post-processing steps to reconnect the resulting tracklets into longer tracks \cite{mansourian2023multi, maglo2023individual}. In our work, we follow a unified, hierarchical approach proposed by Cetintas et al.\cite{Cetintas_2023_CVPR} that addresses both short-term and long-term tracking.  We argue that we need to incorporate domain-specific features such as jersey numbers, team ID, and field position to achieve reliable long-term tracking of players. Experiments on SoccerNet\cite{cioppa2022soccernet} player tracking dataset show that our approach is effective.

To further evaluate the performance of our method we propose a new hockey player tracking dataset. Recorded with a stationary camera, the video clips capture the whole playing surface and have longer video sequences (the average length is 1311 frames). The dataset is particularly challenging for tracking because individual player's features are hidden by protective gear, and motion blur and frequent occlusion make jersey numbers hard to recognize.  The dataset is available at \href{https://github.com/mkoshkina/sports-SUSHI}{https://github.com/mkoshkina/sports-SUSHI}. 

Our main contributions are as follows:
\begin{itemize}
\item We present a new player tracking method, SportsSUSHI built on top of the hierarchical graph-based method SUSHI\cite{Cetintas_2023_CVPR} that allows for the long-term association between tracklets. We show how incorporating domain-specific features such as jersey numbers improves the performance of player tracking. 

\item We propose a novel hockey tracking dataset that is unique in that it captures the whole playing surface with a stationary camera. 

\item To showcase its high performance we evaluate SportsSUSHI our novel hockey dataset and on publicly-available SoccerNet\cite{cioppa2022soccernet}.
\end{itemize}

\section{Related Work}

\subsection{MOT}
In recent years, convolutional networks have driven the success of tracking-by-detection for multi-object tracking (MOT). This approach uses object detectors to identify objects in each frame, followed by data association to link detections to tracks. Challenges such as occlusions, missed detections, and similar object appearances are addressed by tracking "lost" tracks and employing re-identification.

MOT methods are classified as online or offline \cite{ciaparrone2020deep}. Online tracking processes frames sequentially, enabling real-time applications but limiting its ability to correct past errors. Offline tracking, in contrast, processes entire sequences, leveraging both past and future frames for globally optimal trajectories, making it more accurate but unsuitable for real-time use.

Online tracking performs well for short-term tracking and pedestrian datasets, where occlusions and absences are minimal. However, it struggles with player tracking, which involves more complex dynamics and prolonged occlusions.

The majority of the proposed generic MOT methods are online such as  Tracktor++\cite{bergmann2019tracking}, CenterTrack\cite{zhou2020tracking}, FairMOT\cite{zhang2021fairmot}, ByteTrack\cite{zhang2022bytetrack}, SORT\cite{Bewley2016_sort} and DeepSORT\cite{wojke2017simple}. Some of these piggyback on object detectors repurposing them to perform detection and association between a pair of frames in a single step (Tracktor\cite{bergmann2019tracking}, CenterTrack\cite{zhou2020tracking}). The data association approach is often performed using bi-partite matching (with Hungarian algorithm), where matching is sometimes done in stages \cite{wojke2017simple, zhang2022bytetrack}.

The offline MOT problem is often modelled as a graph where nodes represent detections or
tracklets and edges between two nodes are potential associations. One trajectory corresponds to one flow path in the graph. Several early methods were proposed to use this model and solve for an optimal global solution using min-cost network flow algorithms  \cite{zhang2008global, choi2012unified}. The min-cost flow formulation is guaranteed to find an optimal solution in polynomial time.
MPNTrack \cite{braso2020learning} built on this approach but used a graph neural network (GNN) and treated the problem as edge classification. Building the graph based on all frames and all detections is, of course, too expensive.  Therefore, they use a batch approach considering 15 frames at a time. That is a limited window to consider for long-term tracking.

Cetintas et al. \cite{Cetintas_2023_CVPR} use the MPNTrack \cite{braso2020learning} GNN approach to develop a unified model
they call SUSHI which is scalable, generalizable, and suitable for both short-term and long-term tracking. Instead of having a different approach and manual feature architecture for short-term and long-term tracking, they break up an image sequence into a hierarchy. They construct a graph with detections as nodes and employ a GNN to classify the edges. The resulting tracklets are then utilized to create a new graph, connecting them into longer tracklets, and so forth. While the multi-stage approach has been employed in other works, SUSHI \cite{Cetintas_2023_CVPR} distinguishes itself by reusing the same architecture and weights across all levels. SUSHI breaks down
the image sequence into multiple levels, resulting in smaller graphs at each level, which enhances the system’s speed and scalability. To avoid manual feature selection for each hierarchy level, each level has its own trained MLP layer to encode features (such as appearance and geometric features). This way the network is trained to select the most relevant features for each hierarchy level. Due to its speed and scalability, SUSHI is a step forward in offline long-term tracking. Weight-sharing and level-specific feature encoding allow for data-driven level-specific feature selection.
 
Due to its ability to capture long-term associations, SUSHI is well-suited for the task of long-term tracking. In our work, we adopt the SUSHI approach and explore domain-specific features that best guide long-term associations. 

\begin{figure*}[h]
  \centering
\includegraphics[width=0.8\linewidth]{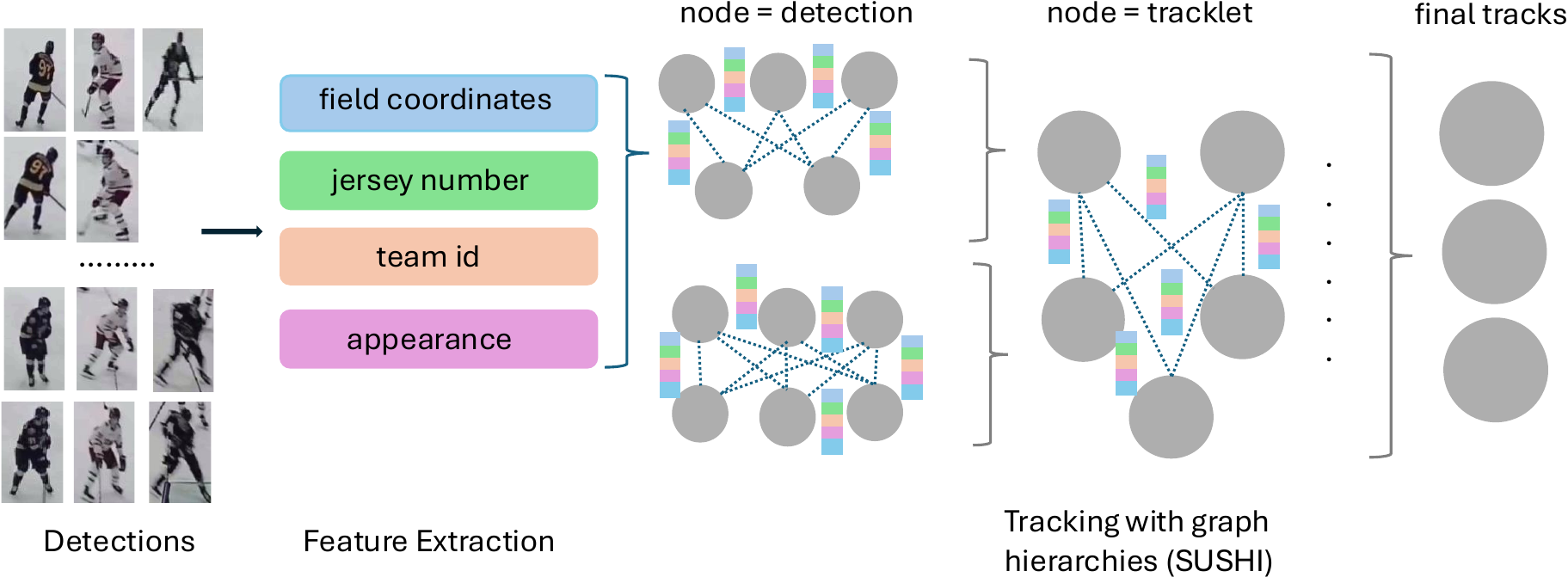}
   \caption{SportsSUSHI: we propose a player tracking system based on hierarchical tracker SUSHI\cite{Cetintas_2023_CVPR}. Our feature extraction module, extracts features crucial for player tracking: jersey number, field coordinates, and team ID, in addition to classic re-ID features. The tracker then builds a hierarchy of graphs where each next level spans longer temporal distances. Initial graphs contain detections as nodes. Similarity measures between node features serve as the edge feature. Each following layer contains tracklets formed by solving the graph in the previous step as nodes.}
    \label{fig:overview}
\end{figure*}

\subsection{Player MOT}
Most Multi-Object Tracking (MOT) methods focus on pedestrians or vehicles, while player tracking (or Multiple Athlete Tracking, MAT) remains less explored, largely due to its niche application and the historical lack of large public datasets. However, the recent introduction of SoccerNet \cite{cioppa2022soccernet} and SportsMOT \cite{cui2023sportsmot} has spurred significant advancements in this field, as evidenced by a rise in research publications \cite{maglo2023individual, cui2023sportsmot, huang2023observation, yang2023hard, mansourian2023multi, lv2024diffmot, huang2024iterative}.

Player tracking leverages both appearance and kinematic features, as in MOT, but requires additional sports-specific features like team IDs and jersey numbers due to the visual similarity among teammates \cite{vats2021player, zhang2020multi, mansourian2023multi}. Unlike pedestrians, players exhibit abrupt, non-linear movements, which tracking methods often address \cite{yang2023hard, huang2023observation, huang2024iterative, lv2024diffmot}. Pose extraction is also commonly used to enhance appearance features and manage occlusions \cite{kong2020online, kong2020long}.

Zhang et al. \cite{zhang2020multi} proposed a multi-camera tracking method incorporating team IDs, jersey numbers, and pose-guided feature extraction into a unified identity feature for track association. Their pipeline, trained separately for each game, demonstrated strong results on the APIDIS\cite{apidis} dataset but required retraining for new games.

Yang et al. \cite{yang2023hard} introduced a Cascaded Buffered IoU (C-BIoU) tracker to handle rapid player movements. This method extends bounding box matching and improves short-term tracking, as seen in its performance on SoccerNet. Similarly, Huang et al. \cite{huang2024iterative} combined ideas from ByteTrack and C-BIoU into Deep-EIoU, achieving superior short-term tracking results.

Lv et al. \cite{lv2024diffmot} addressed motion modelling by replacing the Kalman Filter with a Decoupled Diffusion Model, better suited for abrupt movements in sports. Their method, DiffMOT, achieved state-of-the-art results on SportsMOT but remains focused on short-term tracking.

Two of the most successful methods on SoccerNet dataset \cite{cioppa2022soccernet}, Mansourian et al. \cite{mansourian2023multi} and Maglo et al. \cite{maglo2023individual}, both focus on improving player re-identification and using it to connect tracklets into longer tracks as a post-processing step. Mansourian et al. \cite{mansourian2023multi} use an approach similar to Zhang et al.\cite{zhang2020multi}. Their proposed tracker PRT-Track utilizes keypoint detection to build an appearance feature that is composed of appearance features from different body parts of the player.  The resulting ID feature is less susceptible to occlusions. On the other hand, Maglo et al. \cite{maglo2023individual} approach this by first running an existing short-term tracker, then using resulting tracklets to fine-tune the re-ID system in a self-supervised way to distinguish between players present in the game. They utilize contrastive learning, using detections that belong to the same tracklet as positives and detections that are captured in the same frame but belonging to other tracklets as negatives.  After fine-tuning the re-ID network they use these new re-ID features to connect tracklets into longer tracks.  Maglo et al. \cite{maglo2023individual} reports state-of-the-art results on both SoccerNet\cite{cioppa2022soccernet} and SportsMOT\cite{cui2023sportsmot}. This shows the importance of reliable re-ID feature and its role in long-term tracking.  This approach, although effective, requires a lot of overhead as it includes training on every single test game separately at inference time.

Our approach utilizes domain knowledge similarly to \cite{mansourian2023multi} by extracting jersey number, team id and re-ID feature. However, using hierarchical graphs provides a unified approach to both short-term and long-term tracking and does not require re-training on each new test game as does Maglo et al.\cite{maglo2023individual}.

\subsection{Jersey Number Recognition}
Gerke et al. \cite{gerke2015soccer} and Li et al. \cite{li2018jersey} pioneered CNN-based approaches for jersey number recognition, which remain dominant. Liu et al. \cite{liu2019pose,liu2022jede} further improved classification by integrating body pose detection with Faster R-CNN \cite{ren2015faster} and Mask R-CNN \cite{he2017mask}.

Vats et al. \cite{vats2021multi} showed that multi-task training on both holistic and digit-wise number classification outperforms single-task training. However, their large labelled dataset is not public, lacks all possible jersey numbers, and struggles to generalize. Bhargavi et al. \cite{bhargavi2022knock} addressed this by pre-training on synthetic data and fine-tuning on a small real-world dataset.

Most prior approaches treat jersey number recognition
as a specialized classification problem requiring the design and training of a dedicated classification network. In
contrast, \cite{Koshkina_2024_CVPR, nady2021player, chen2023tracking} use a more generally trained scene text recognition (STR) model that allows handling of all possible jersey numbers, instead of being restricted to numbers
that happen to be in the training dataset. Koshkina et al.\cite{Koshkina_2024_CVPR} propose a pipeline that uses filtering out of players that don't have visible jersey numbers, before relying on pose estimation to detect jersey number on the player and predict the number using PARSeq\cite{bautista2022parseq} model. They report high accuracy on both hockey and soccer datasets. We use their approach as a pre-processing step to estimate jersey numbers when they are visible.

\subsection{Team Identification}
Early methods for team affiliation classification relied on color histograms \cite{mazzeo2010football,shtrit2011,ivankovic2014automatic,d2009investigation,lu2013learning,liu2014detecting,bialkowski2014representing} and "bag of words" representations of color features \cite{tong2011automatic}. While lightweight, these approaches are sensitive to illumination changes and perform poorly when teams wear similar colors.

More recent supervised deep learning methods \cite{lu2018lightweight, istasse2019associative} improve performance but require labelled data. Istasse et al. \cite{istasse2019associative} trained a CNN to segment players and generate pixel-wise team descriptors, clustering these to identify teams. However, this approach requires pixel-level labels for training and does not provide instance-level segmentation, limiting its use for applications like player location heatmaps.

Lu et al. \cite{lu2018lightweight} used a cascaded CNN for team classification (team A, team B, and others) but required fine-tuning on labelled data for each new game, limiting generalization.

Koshkina et al.\cite{koshkina2021contrastive} adopted a contrastive learning approach in which an embedding network learns to maximize the distance between representations of players on different teams relative to players
on the same team. Referees whose uniforms are typically consistent between games are identified using a supervised classification network.  In our work, we use this method of team identification in our feature extraction phase.

\subsection{Field Registration}
In broadcast videos, the camera movement makes the position of the player in the frame an unreliable feature. Given the importance of spatial information for tracking, establishing the player's position that is independent of the span and zoom of the camera is critical. Sports field registration methods can obtain a homography from a given camera view to a 2D field surface position. Several proposed methods use convolutional neural networks or transformers to find key points and line segments in the frame \cite{gutierrez2024no, maglo2023individual, theiner2023tvcalib}. Typically, the RANSAC method is used to estimate the homography matrix. In our work, we use the system proposed by Guti{\'e}rrez-P{\'e}rez et al. \cite{gutierrez2024no} to obtain frame-to-field homography. We then estimate player positions on the playing surface.


\section{Method}
Our tracking-by-detection approach begins by taking the detected bounding boxes of players and processing them through a feature-extraction module. This module predicts key attributes such as jersey numbers, field coordinates, team IDs, and appearance features (Figure~\ref{fig:overview}). Tracking is performed offline by recursively constructing a hierarchy of graphs.  At the initial level, the nodes in the graph represent detections, and edges represent potential associations between detections across consecutive frames. The edge features are derived from the distance between detection features. Solving these initial graphs produces short tracklets.  In subsequent levels, the tracklets from the previous step become the nodes of new graphs, with edges representing associations between tracklets over longer time spans. This hierarchical process is repeated for N levels, enabling the system to handle increasingly longer temporal relationships.  For video clips exceeding the longest temporal distance considered, the method processes them using a sliding window approach, applying a simple stitching scheme, as described in the original method \cite{cetintas2023unifying}.

\subsection{SportsSUSHI}
We extend hierarchical graph tracker SUSHI\cite{cetintas2023unifying} to support domain-specific features.  The strength of SUSHI lies not only in considering long-term connections but in having an architecture that on one hand shares GNN weights between different levels but on the other learns separate feature encoders for each level. This allows the network to learn which features are more important at each temporal distance. At each level, once edge features are encoded using MLP, the GNN is solved for the task of binary edge classification using neural message passing\cite{braso2020learning}. The linear program is then used to convert edge predictions into binary decisions and obtain final trajectories. For more information on SUSHI, refer to the original paper \cite{cetintas2023unifying}.

In the original SUSHI implementation, the authors use spatial features, such as the position and size of the bounding box, as well as the appearance (re-ID) feature. Unfortunately, these features are not sufficient for accurate player tracking.  Based on our experiments we found that using sports-domain features improves tracking performance. Therefore, we introduce jersey number prediction and team ID to supplement appearance features. In broadcast videos, where the camera follows the play, comparing positions of bounding boxes between frames becomes unreliable, especially at longer intervals. To mitigate this, we use field registration as part of the pre-processing stage to estimate the homography matrix between the image and 2D playing field map. We then use projected player positions on the field as the spatial feature. In the following subsections, we discuss the feature extraction process in more detail. 

\begin{figure*}[h]
  \centering
\includegraphics[width=0.85\linewidth]{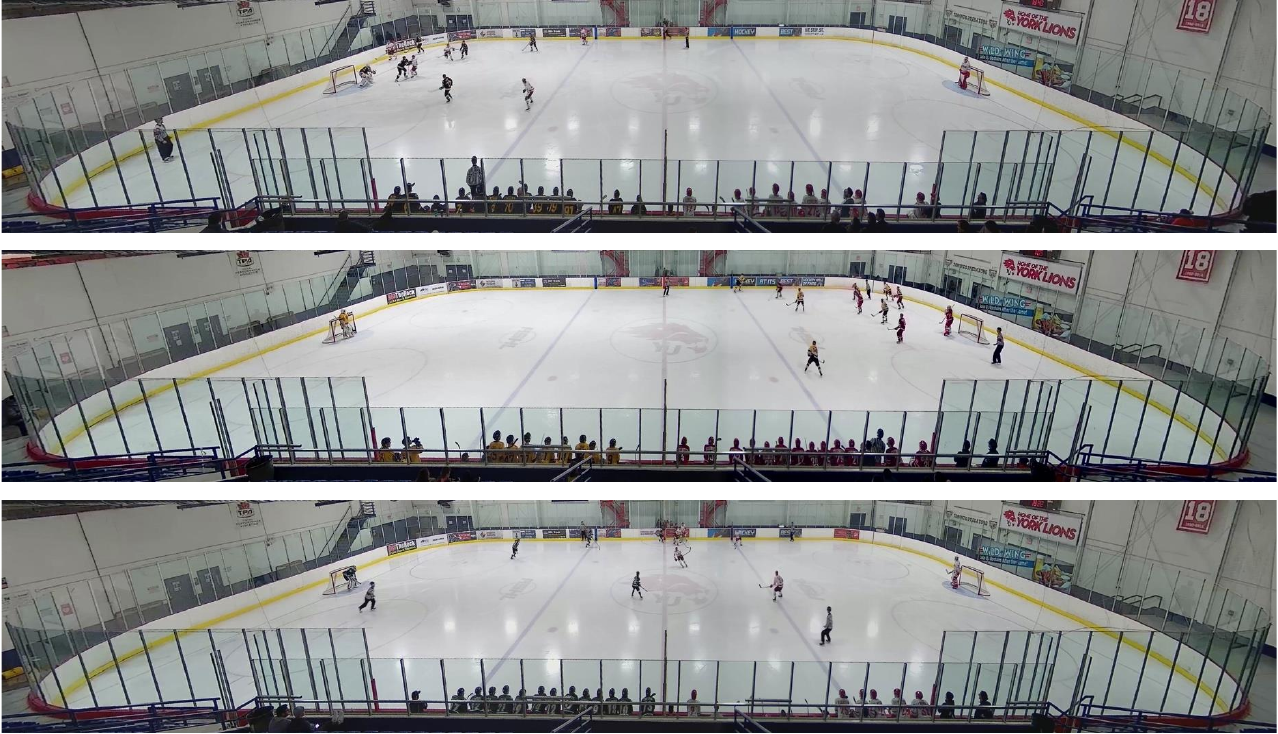}
   \caption{Sample frames from our hockey dataset.}
    \label{fig:hockey_dataset_samples}
\end{figure*}

\subsection{Feature Extraction}
\subsubsection{Jersey Number Prediction} 
We make use of the jersey number recognition pipeline introduced by Koshkina et al. \cite{Koshkina_2024_CVPR}. The pipeline takes player detections as input and classifies them to identify player images with legible jersey numbers. For these player images, pose estimation is used to select the torso area containing the number, before passing it to the scene text recognition network (STR) PARSeq\cite{bautista2022parseq}. Before decoding its prediction into a character string, PARSeq produces a confidence vector for each possible character value at each position. For a jersey number, $c_1=[c_1(EOL), c_1(0), c_1(1) ... c_1(9)]$ and $c_2=[c_2(EOL), c_2(0), c_2(1) ... c_2(9)]$ where $EOL$ is an end-of-line character and $c_i(d)$ is a confidence of having digit $d$ in position $i$. We mark any image with predictions of $EOL$ in the first position as illegible. For the rest, we calculate $c(d_id_j) = c_1(d_i)c_2(d_j)$ for two-digit numbers and $c(d_i) = c_1(d_i)c_2(EOL)$ for one-digit numbers. The resulting 100-value vector encodes jersey number prediction.

\subsubsection{Team ID Prediction}
Our experiments show that the appearance feature is not sufficient for some datasets to distinguish between players on different teams. To address that, we also include the team ID feature. The team ID feature in our system is a one-hot vector that represents a prediction of whether the player belongs to team A, team B, or is a referee. We use a method proposed by Koshkina et al.\cite{koshkina2021contrastive} to first classify a person as a referee or a player, then extract the image embedding for each player using a previously trained embedding network.  Both the referee classifier and the player embedding networks are trained on the training set. At inference, each detection is classified as player or referee first. For each test clip, the first N player images are clustered to learn Team A and Team B's appearance for the given clip. All of the players in the clip are then assigned a label based on the closest cluster center.  The team ID feature is therefore a one-hot vector of length 3.

\subsubsection{Field Coordinates}
We make use of field registration work by Guti{\'e}rrez-P{\'e}rez et al.\cite{gutierrez2024no}. They use convolutional network encoder/decoder architecture to detect key points and line segments on the field and predict each frame's frame-to-field homography matrix. After learning each frame's homography matrix we use it to project a middle point of each detection bounding box in the frame (in pixels) to a position on a field (in meters).  Field registration sometimes fails to detect key points in the frame. This typically happens when a large zoom and camera angle don't capture enough field markings or when the camera motion results in the blurring of lines. To address this, we use linear interpolation of predicted camera parameters to approximate parameters for failed frames. 

\subsubsection{Person Re-ID}
For some sports, for example, soccer visual appearance provides a lot of useful information despite similar uniforms of players on the same team. As in the original SUSHI method, we extract the appearance feature vector using a person re-identification model \cite{he2020fastreid}. We fine-tune the model on a specific dataset beforehand.

\subsubsection{Edge Features}
 For each edge connecting two detections (or two tracklets) we calculate the edge feature vector. Each value of the vector corresponds to a measure of similarity between these two nodes in the given feature space (re-ID, jersey number, etc). We use cosine similarity for re-ID, jersey number, team ID features, and absolute value for distance in meters of field coordinates. For nodes representing tracklets appearance features are computed by taking a mean of individual detection values. For position, the distance between the last detection of the first tracklet and the first detection of the second tracklet is computed. Note, that extracted features for each detection are used only to calculate the edge features. MLP (one for each hierarchy level) is then used to encode the edge features. To control the complexity of the graph Cetintas et al. \cite{cetintas2023unifying} employ the edge-pruning method to only consider K closest nodes based on appearance and spatial features. We follow this approach with K=10.  The resulting graph is used as input into GNN and is solved with neural message passing.

\begin{table*}[h]
  \centering
  {\small{
  \begin{tabular}{lllllll}
  \\
  \toprule
    Dataset & Train & Test & Avg Length (frames) & Frame Rate (fps) & Dimensions (pixels)\\
    \midrule
    Hockey (ours) & 14 & 6 & 1,131 & 30 & 5930 x 1080\\
    SoccerNet\cite{cioppa2022soccernet} & 57 & 49 & 750 & 25 & 1920 x 1080 \\
    \bottomrule
  \end{tabular}
  }}
  \caption{Dataset information.}
  \label{tab:datasets}
\end{table*}

\section{Datasets}
There are several player tracking datasets that became available in the last few years: SoccerNet\cite{cioppa2022soccernet}, SportsMOT\cite{cui2023sportsmot}, and MHPTD\cite{MHPTD}. All of these contain clips from broadcast videos. To explore longer-term tracking we introduce a new hockey tracking dataset.  The dataset is unique in capturing the whole playing surface.  We evaluate SportsSUSHI on our hockey dataset and publicly-available SoccerNet\cite{cioppa2022soccernet} tracking dataset.

\subsection{Hockey}
\begin{figure}[t]
  \centering
   \includegraphics[width=1.0\linewidth]{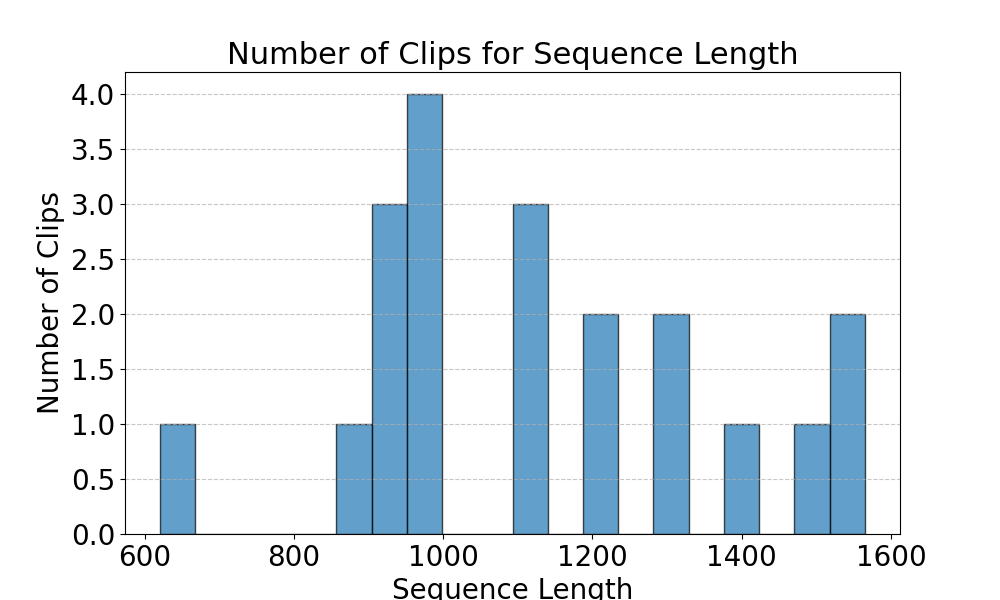}

   \caption{Hockey dataset sequence length.}
   \label{fig:seq_length}
\end{figure}

Recorded at university hockey games with a stationary camera at 30fps, the dataset contains 20 clips from 9 different games. We split it into 14 training set and 6 test set clips, keeping clips from the same game in the same partition. The average clip length is 1311 and the longest is 1530 frames (Figure~\ref{fig:seq_length} shows sequence length distribution). 
The dataset contains ground truth in standard MOT annotation. In most MOT datasets when the person is occluded there are no bounding box annotations for them. Motivated to investigate tracking a person throughout the time they are on the playing surface, we estimated bounding boxes of occluded players and included these in the annotation. 

The annotation process was a two-step approach:
\begin{itemize}
    \item Weak tracking ground truth was generated by annotators following each player in turn with a mouse pointer throughout the clip. Bounding boxes from an object detector were then associated with each such path, producing a set of imperfect tracks.
    \item These tracks were then manually corrected and refined to ensure accuracy. Any missing bounding boxes were added and boundaries were adjusted. We used the CVAT \cite{CVAT_ai_Corporation_Computer_Vision_Annotation_2023} annotation tool for this step.  During this stage, jersey numbers (whenever visible) and team ID labels were added to the annotation. 
\end{itemize} 
Figure~\ref{fig:hockey_dataset_samples} shows sample frames for our new hockey dataset. Table~\ref{tab:datasets} summarizes some statistics of the two datasets used.

\subsection{SoccerNet}
SoccerNet\cite{cioppa2022soccernet} player tracking dataset introduced in 2022 is the first large-scale publicly available player tracking dataset. Its introduction and the corresponding challenge spurred interest in the area where in the past the lack of publicly available data posed a barrier to researchers. SoccerNet contains 200 clips, spread into train, test and challenge sets. The clips are from broadcast soccer games and contain a lot of challenging scenarios: camera motion, fast player movement, occlusions, and players leaving and re-entering a field of view. The last condition is especially challenging and relies heavily on the ability of the tracker to re-identify the player and reconnect parts of the track after long intervals.

\begin{table*}[h]
  \centering
  {\small{
  \begin{tabular}{llll} 
  \toprule
    Method & HOTA & AssA & DetA \\
    \midrule
     & \multicolumn{3}{c}{Ground Truth Detections}    \\
    \midrule
    SUSHI (9 layers) \cite{cetintas2023unifying} & 85.79 & 75.55 & 97.42 \\
    SUSHI (ft re-ID, 9 layers) & 85.37 & 79.36  & 92.20 \\
    SportsSUSHI (ft reid, field position, 9 layers) & 89.22 & 84.25 & 94.82 \\
    SportsSUSHI (ft reid, field position, jersey numbers, 7 layers) & 87.44 & 80.92 & 94.87 \\
    SportsSUSHI (ft reid, field position, jersey numbers, 9 layers) & 89.78 & 85.35  & 94.80 \\ 
    SportsSUSHI (ft reid, field position, jersey numbers, 10 layers) & \textbf{90.92} & \textbf{87.51} & 94.80\\
    \bottomrule
  \end{tabular}
  }}
  \caption{Abblations on SoccerNet \cite{cetintas2023unifying} test partition.}
  \label{tab:SportsSUSHI-abblations}
\end{table*}

\begin{table}[h]
  \centering
  {\small{
  \begin{tabular}{lllll}
  \toprule
      & \multicolumn{2}{c}{Soccer} & \multicolumn{2}{c}{Hockey}\\
     \midrule
    Step (frames) & Original & Soccer & Original & Hockey \\
     \midrule    
    1 & 98.0 & 99.1 & 99.1 & 99.63 \\
    50 & 51.2 & 79.1 & 32.5 & 38.2 \\
    100 & 35.7 & 72.2 & 25.5 & 26.8 \\
    300 & 33.0 & 62.2 & 29.9 & 26.6 \\
    \bottomrule
  \end{tabular}
  }}
  \caption{FastReID performance analysis for matching player identities at varying frame intervals using cosine similarity of feature vectors. Accuracy is compared between the Market1501-trained model and its fine-tuned version for each dataset.}
  \label{tab:reid}
\end{table}

\section{Experiments and Results} 
\subsection{Re-ID Performance Analysis}

For appearance feature extraction, we use the FastReID model with ResNet50-IBN backbone \cite{he2020fastreid} originally trained on the Market1501 dataset \cite{zheng2015scalable} and finetuned on the tracking dataset. We investigate how reliable the similarity of re-ID features is at different time gaps. To do that we take a subset of the test set and analyze the accuracy of matching the identity of the player in frame $i$ to that same player in frame $i+N$ for different $N$. We can see from the results in Table~\ref{tab:reid} that re-identification does predictably worse at large time intervals. We can also observe that for soccer where the player's distinctive features such as face, skin colour, hair colour and style are visible, person re-identification performs better than for hockey. These findings motivate using jersey ID as a feature for both datasets and team identification for hockey.

\subsection{Implementation}
For both jersey number recognition framework, team ID, and field registration we use the code and model weights from the original papers (\cite{Koshkina_2024_CVPR}, \cite{koshkina2021contrastive}, and \cite{gutierrez2024no} accordingly)

Our SportsSUSHI method uses the original SUSHI implementation but replaces feature extraction and edge feature calculation components. We train SportsSUSHI on the training partition of the player tracking dataset following the training protocol outlined in Cetintas et al. \cite{cetintas2023unifying}: training earlier levels first for 500 iterations before unfreezing each next level. We then continue to train jointly for a total of 250 epochs. The model is trained with the Adam optimizer.

We evaluate tracking results using key standard HOTA\cite{luiten2021hota} metrics: Detection Accuracy, Association Accuracy, and HOTA. Since our goal is reliable long-term tracking we are most interested in Association Accuracy and HOTA metrics.
\begin{table}[h]
  \centering
  {\small{
  \begin{tabular}{llll} 
  \toprule
    Method & HOTA & AssA & DetA \\
    \midrule
     & \multicolumn{3}{c}{Ground Truth Detections}    \\
    \midrule
    Maglo et al. \cite{maglo2023individual}* & \textbf{96.57} & \textbf{93.60} & 99.65 \\
    Mansourian et al. \cite{mansourian2023multi}* & 90.77 & 82.53 & 99.83 \\
    Yang et al. \cite{yang2023hard}* & 89.2 & 80.00 & 99.4 \\
    Huang et al. \cite{Huang_2023_WACV}* & 85.44 & 73.57 & 99.24 \\
    SUSHI \cite{cetintas2023unifying} & 85.79 & 75.55 & 97.42 \\
    \midrule
    \textbf{SportsSUSHI (ours)} & \textbf{90.92} & \textbf{87.51} & 94.80\\
    \midrule
    & \multicolumn{3}{c}{YOLOX Detections}  \\
     \midrule    
    Maglo et al. \cite{maglo2023individual}* &  \textbf{73.29} & \textbf{73.42} & 73.26\\
    Mansourian et al. \cite{mansourian2023multi} * & 59.77 & 58.55 & 61.09\\
    ByteTrack \cite{zhang2022bytetrack} & 60.56 & 52.45 & 70.10 \\
    \midrule
    \textbf{SportsSUSHI (ours)} & \textbf{71.36} & \textbf{69.99} & 72.87\\ 
    \bottomrule
  \end{tabular}
  }}
  \caption{Results on SoccerNet \cite{cetintas2023unifying} test partition. (results marked with * are taken from the corresponding paper)}
  \label{tab:soccernet-results}
\end{table}

\begin{table}
  \centering
  {\small{
  \begin{tabular}{llll} 
  \toprule
    Method & HOTA & AssA & DetA \\
     \midrule    
    Tracktor++ \cite{bergmann2019tracking} & 44.33  & 34.09& 58.08\\ 
    CenterTrack \cite{zhou2020tracking} & 59.70 &  48.96 & 72.90\\
    FairMOT \cite{zhang2021fairmot} & 61.56 & 55.51 & 68.35 \\
    ByteTrack \cite{zhang2022bytetrack} & 67.51 & 64.63 & 71.36\\
    \midrule
    \textbf{SportsSUSHI (ours)} & \textbf{71.24} & \textbf{72.82} & 70.47 \\    
    \bottomrule
  \end{tabular}
  }}
  \caption{Results on hockey dataset test partition.}
  \label{tab:hockey-results}
\end{table}

\subsection{Soccer}
The best results on the SoccerNet dataset were achieved by combining field coordinates, re-ID appearance, and jersey number features. Our SportsSUSHI tracker shows competitive performance compared to earlier published methods. Table ~\ref{tab:soccernet-results} presents the results on both ground truth detections and YOLOX detections \cite{ge2021yoloxexceedingyoloseries}.

We show the contribution of various features to performance in Table~\ref{tab:SportsSUSHI-abblations}. We use the SUSHI out-of-the-box model without any fine-tuning on the dataset as our baseline. We show that using field coordinates instead of frame pixel coordinates introduces a significant performance improvement. The introduction of jersey numbers provides a further performance boost.

We explore the effect of the number of layers in the graph hierarchy. With 7 layers the furthest time gap in a track considered for reconnection is 256, for 9 layers it is 512 and for 10 layers it is 1024. The performance improves with additional layers since the model gets to consider longer timespans (Note, the total length of SoccerNet clips is 750 frames, so further layers are not needed).

\subsection{Hockey}
In our hockey dataset, the camera is stationary and captures the whole rink. Therefore, using the original spatial feature (position of the player in the frame and bounding box size) is sufficient and we don't need to calculate the player's position on the rink. We use the jersey number, team ID, and re-ID features in addition to spatial ones. We show results on our test set in Table~\ref{tab:hockey-results}. All of the baselines we compare with have been fine-tuned on the hockey dataset. We use YOLOX\cite{ge2021yoloxexceedingyoloseries} detections in our method. SportsSUSHI shows superior results compared to previous methods.

\subsection{Failure Cases}
Using longer timestamps to connect tracklets and employing jersey IDs for re-identifying players yield promising results in player tracking. However, jersey numbers are not always visible or are often too blurry to decipher. Moreover, they offer no additional cues when tracking referees, who all wear the same uniform and are difficult to re-identify. The association errors observed in the hockey and soccer datasets primarily arise from failures to re-identify individuals after extended absences from the field of view or during occlusions. A potential next step could involve incorporating a motion model to predict player movements, which would enhance the utility of spatial features when appearance-based features are insufficient. We leave this to future work.

\section{Conclusion}
In this paper, we proposed SportsSUSHI, a hierarchical graph-based approach designed for long-term player tracking in team sports. By incorporating domain-specific features such as jersey numbers, team IDs, and field coordinates, SportsSUSHI addresses key challenges like occlusion, camera movement, long absences of players from the field of view, and player similarity within teams. These features significantly enhance the system's ability to maintain accurate associations across occlusions and rapid player movements. Evaluations conducted on the SoccerNet dataset and a newly introduced hockey dataset demonstrate the method's effectiveness, particularly in improving association accuracy. 

In the future, we plan to investigate incorporating a motion model that can predict both past and future player positions, addressing the complex dynamics of sports motion patterns. Additionally, we look to expand the method's applicability to diverse sports scenarios to further improve its robustness and versatility.

\clearpage
{\small
\bibliographystyle{ieee_fullname}
\bibliography{player_tracking}
}

\end{document}